\documentclass{article}
\usepackage{ifpdf}

\usepackage{hyperref}       

\usepackage{arxiv}
\usepackage[ruled,vlined]{algorithm2e}
\usepackage{float}
\usepackage{multirow}
\usepackage{booktabs}
\usepackage{bm}

\usepackage[utf8]{inputenc} 
\usepackage[T1]{fontenc}    

\usepackage{url}            
\usepackage{booktabs}       
\usepackage{amsfonts}       
\usepackage{nicefrac}       
\usepackage{microtype}      
\usepackage{lipsum}
\usepackage{graphicx}

\title{CrisisBERT: A Robust Transformer for Crisis Classification and Contextual Crisis Embedding}

\author{
 Junhua Liu$^{1,2}$, Trisha Singhal$^1$, Lucienne T.M. Blessing$^1$, Kristin L. Wood$^1$, Kwan Hui Lim$^1$ \\\\
 \texttt{j@forth.ai, \{trisha\_singhal, lucienne\_blessing, kristinwood, kwanhui\_lim\}@sutd.edu.sg} \\\\
  $^1$Singapore University of Technology and Design\\
  $^2$ProQod Singapore \\
}

\begin{document}
\maketitle
\begin{abstract}
Classification of crisis events, such as natural disasters, terrorist attacks and pandemics, is a crucial task to create early signals and inform relevant parties for spontaneous actions to reduce overall damage. Despite the crises, such as natural disaster, that can be predicted by professional institutions, certain events are first signaled by everyday citizens, i.e., civilians, such as the recent COVID-19 pandemics. Social media platforms such as Twitter often expose firsthand signals on such crises through high volume information exchange. In the case of Twitter, this corresponds to on average over half a billion tweets posted daily. Prior works proposed various crisis embeddings and classification using conventional Machine Learning and Neural Network models. However, none of the works perform crisis embedding and classification using state of the art attention-based deep neural networks models, such as Transformers and document-level contextual embeddings. This work proposes \textit{CrisisBERT}, an end-to-end transformer-based model for two crisis classification tasks, namely crisis detection and crisis recognition, which shows promising results across accuracy and f1 scores. The proposed model demonstrates superior robustness over various benchmarks, as it shows marginal performance compromise while extending from 6 to 36 events with only 51.4\% additional data points. We also propose \textit{Crisis2Vec}, an attention-based, document-level contextual embedding architecture for crisis embedding, which achieves better performance than conventional crisis embedding methods such as \textit{Word2Vec} and \textit{GloVe}. To the best of our knowledge, our works are first to propose using transformer-based crisis classification and document-level contextual crisis embedding in the literature.
\end{abstract}


\section{Introduction}

Crisis-related events, such as earthquakes, hurricanes and train or airliner accidents, often stimulate a sudden surge of attention and actions from both media and the general public. Despite the fact that crises, such as natural disasters, can be predicted by professional institutions, certain events are first signaled by everyday citizens, i.e., civilians. For instance, the recent COVID-19 pandemics was first informed by general public in China via \textit{Weibo}, a popular social media site, before pronouncements by government officials. 

Social media sites have become centralized hubs that facilitate timely information exchange across government agencies, enterprises, working professionals and the general public. As one of the most popular social media sites, Twitter enables users to asynchronously communicate and exchange information with \textit{tweets}, which are mini-blog posts limited to 280 characters. There are on average over half a billion tweets posted daily\cite{twitter_stats}. Therefore, one can leverage on such high volume and frequent information exchange to expose firsthand signals on crisis-related events for early detection and warning systems to reduce overall damage and negative impacts. 

Event detection from tweets has received significant attention in research in order to analyze crisis-related messages for better disaster management and increasing situational awareness. Several recent works studied various natural crisis events, such as hurricanes and earthquakes, and artificial disasters, such as terrorist attacks and explosions\cite{said2019natural, snyder2019situational, imran2015processing, sakaki2010earthquake}.

These works focus on binary classifications for various attributes of crisis, such as classifying source type, predicting relatedness between tweets and the crises, and assessing informativeness and applicability~\cite{olteanu2014crisislex, zhang2016semi, stowe2016identifying}. On the other hand, several works proposed multi-label classifiers on affected individuals, infrastructure, casualties, donations, caution, advice, etc.~\cite{imran2013extracting, hughes2014online}. Crisis recognition tasks are likewise conducted such as identifying crisis types, i.e. hurricanes, floods and fires~\cite{burel2017semantics, crow2020verifying}. 

Machine Learning-based models are commonly introduced in performing the above mentioned tasks. Conventional linear models such as Logistic Regression, Naive Bayes and Support Vector Machine (SVM) are reported for automatic binary classification on informativeness ~\cite{parilla2014automatic} and relevancy~\cite{stowe2016identifying}, among others. These models were implemented with pre-trained word2vec embeddings~\cite{mikolov2013distributed}. Several unsupervised approaches are also proposed for classifying crisis-related events, such as the CLUSTOP algorithm utilizing Community Detection for automatic topic modelling~\cite{lim2017clustop}. A transfer-learning approach is also proposed~\cite{pedrood2018mining}, though its classification is only limited to two classes. The ability for cross-crisis evaluation remains questionable.

More recently, numerous works proposed Neural Networks (NN) models for crisis-related data detection and classification. For instance, ALRashdi and O’Keefe investigated two deep learning architectures, namely Bidirectional Long Short-Term Memory (BiLSTM) and Convolutional Neural Networks (CNN) using domain-specific and GloVe embeddings~\cite{alrashdi2019deep}. Nguyen \textit{et al.} propose a CNN-based classifier with Word2Vec embedding pretrained on Google News~\cite{mikolov2013distributed} and domain-specific embeddings~\cite{nguyen2017robust}. Lastly, parallel CNN architecture was proposed to detect disaster-related events using tweets~\cite{kim2014convolutional, kersten2019robust}.

While prior works report remarkable performance on various crisis classification tasks using NN models and word embeddings, no studies are found to leverage the most recent Natural Language Understanding (NLU) techniques, such as attention-based deep classification models~\cite{vaswani2017attention} and document-level contextual embeddings~\cite{reimers2019sentence}, which reportedly improve state-of-the-art performance for many challenging natural language problems from upstream tasks such as Named Entity Recognition and Part of Speech Tagging, to downstream tasks such as Machine Translation and Neural Conversation. 


\subsection{Main Contributions}

This work focuses on deep attention-based classification models and document-level contextual representation models to address two important crisis classification tasks. We study recent NLU models and techniques that reportedly demonstrated drastic improvement on state-of-the-art and localize for domain-specific crisis related tasks. 

Overall, our main contribution of this work includes:

\begin{itemize}  
\item proposing \textit{CrisisBERT}, an attention-based classifier that improves state-of-the-art performance for both crisis detection and recognition tasks;
\item demonstrating superior robustness over various benchmarks, where extending \textit{CrisisBERT} from 6 to 36 events with 51.4\% of additional data points only results in marginal performance decline, while increasing crisis case classification by 500\%; 
\item proposing \textit{Crisis2Vec}, a document-level contextual embedding approach for crisis representation, and showing substantial improvement over conventional crisis embedding methods such as \textit{Word2vec} and \textit{GloVe}
\ldots
\end{itemize} 

To the best of our knowledge, this work is the first to propose a transformer-based classifier for crisis classification tasks. We are also first to propose a document-level contextual crisis embedding approach.

\begin{figure*}[t]
  \includegraphics[width=\textwidth]{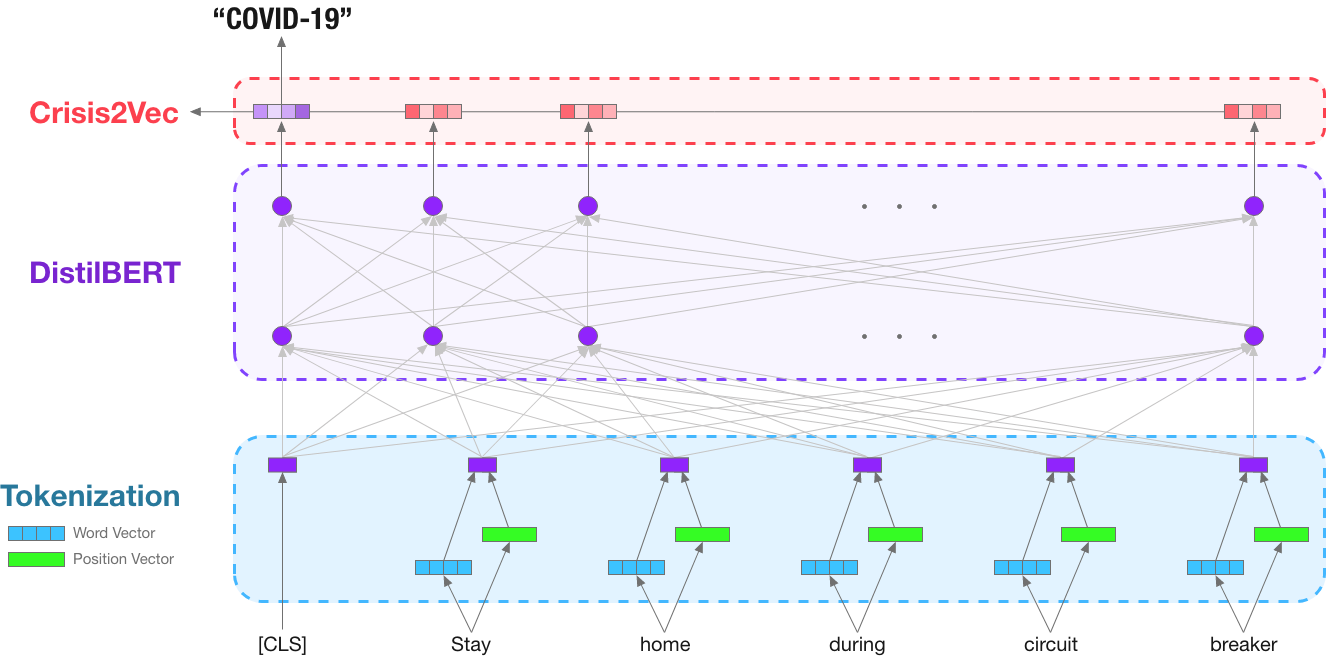}
  \centering
  \caption{Illustration of \textit{CrisisBERT} and \textit{Crisis2Vec} models. During the tokenization phase, word-level and subword-level embeddings are created to produce context embedding for each input word. The vectors are then prefixed with a class header (shown as \textit{[CLS]} in the figure) and passed into a DistrilBERT model. Since we are performing classification task, the CLS token vector, i.e. the first output vector, is then passed into a linear classifier for detection or recognition task, whereas the remainder of the output vectors are average-pooled to create \textit{Crisis2Vec} embeddings.}
  \label{fig:model}
\end{figure*}

\section{Attention-based Neural Crisis Classifier}

In this section, we discuss the recent works that propose various machine learning approaches for crisis classification tasks. While these works report substantial improvement in performance over prior works, none of the works uses state of the art attention-based models, i.e., Transformers~\cite{vaswani2017attention}, to perform crisis classification tasks. We propose \textit{CrisisBERT}, a transformer-based architecture that builds upon a Distilled BERT model, fine-tuned by large-scale hyper-parameter search. 

\subsection{Conventional Crisis classifiers}

Various works propose linear classifiers for crisis-related events. For instance, Parilla-Ferrer \textit{et al.} proposed an automatic binary classification, based on informative and uninformative tweets using Naive Bayes and Support Vector Machine (SVM)\cite{parilla2014automatic}. A SVM with pretrained word2vec embeddings approach was also proposed~\cite{mikolov2013distributed}. 

Besides linear models, recent works also propose deep learning based methods with different neural network architectures. For instance, ALRashdi and O’Keefe investigated Bidirectional Long Short-Term Memory (BiLSTM) and Convolutional Neural Network (CNN) models using domain-specific and GloVe embeddings~\cite{pennington2014glove}. Nguyen \textit{et al.} proposed a CNN model to classify tweets to get information types using Google News and domain-specific embeddings~\cite{nguyen2017robust}.

\subsection{Transformer}

In 2017, Vaswani \textit{et al.} from Google introduced Transformer~\cite{vaswani2017attention}, a new category of deep learning models which are solely attention-based and without convolution and recurrent mechanisms. Later, Google proposed the Bidirectional Encoder Representations from Transformers (BERT) model~\cite{devlin-etal-2019-bert} which drastically improved state-of-the-art performance for multiple challenging Natural Language Processing (NLP) tasks. Since then, multiple transformer-based models have been introduced, such as GPT~\cite{radford2019language} and XLNet~\cite{yang2019xlnet}, among others. Transformer-based models were also deployed to solve domain specific tasks, such as medical text inferencing~\cite{lee2019ncuee} and occupational title embedding~\cite{liu2019ipod}, and demonstrated remarkable performance. 

The Bidirectional Encoder Representation of Transformer (BERT), for instance, is a multi-layer bidirectional Transformer en-
coder with attention mechanism~\cite{devlin-etal-2019-bert}. The proposed BERT model has two variants, namely (a) BERT Base, which has 12 transformer layers, a hidden size of 768, 12 attention heads, and 110M total parameters; and (b) BERT Large, which has 24 transformer layers, a hidden size of 1024, 16 attention heads, and 340M total parameters. BERT is pre-trained with self-supervised approaches, i.e., Masked Language Modeling (MLM) and Next Sentence Prediction (NSP).

While Transformers such as BERT are reported to perform well in natural language processing, understanding and inference tasks, to the best of our knowledge, no prior works propose and examine the performance of transformer-based models for crisis classification. 

\subsection{CrisisBERT} \label{sec:CrisisBERT}

In this work, we investigate the transformer approach for crisis classification tasks and propose \textit{CrisisBERT}, a transformer-based classification model that surpasses conventional linear and deep learning models in performance and robustness. Figure~\ref{fig:model} illustrates an overview of the proposed architecture. Overall, the architecture of \textit{CrisisBERT} includes three layers, namely (1) Contextual Tokenization, (2) Transformer Language Model, and (3) Logistic Classifier. 

\textbf{Tokenization}. DistrilBERT inherits the embedding layer from BERT, where each word is tokenized contextually with subword-level word embeddings and positional encoders. Prefixed with a special token ([CLS]), the vector pairs are concatinated and passed into a DistilBERT LM for training.

\begin{table}[h]
\centering
\resizebox{.6\textwidth}{!}{%
\begin{tabular}{ccc}
\hline
Hyper Parameter & Final & Search Space  \\ \hline
Transformer & BERT & \{BERT, XLNet, GPT2, RoBERTa\}  \\ 
Distillation & True & \{True, False\}  \\ 
Optimizer & ADAMW & \{AdamW, Adam, SGD\}  \\ 
Learning Rate & 5e-5 & \{5e-3, 2e-3, 5e-4, 2e-4, 5e-5, 2e-5\}  \\ 
Batch Size & 32 & \{16, 32, 64\}  \\ 
\hline
\end{tabular}}
\caption{Hyper-parameter search space and final values used}
\label{tab:hyperpara}
\end{table}

\textbf{Transformer LM}. We perform large-scale model selection and hyperparameter search to fin the best performing transformer LM for the crisis classification tasks. Several transformer models are investigated, including \textit{BERT}, \textit{XLNet}, \textit{GPT2} and \textit{RoBERTa}. For each model, we conduct hyper-parameter search, where the search space includes variations of distillation~\cite{hinton2015distilling}, optimizers, learning rates, and batch sizes. 

Table~\ref{tab:hyperpara} shows the breakdown of the search space and the final hyper-parameters for CrisisBERT. Each set of parameters is randomly chosen and ran with 3 epochs and two trials. In total, we evaluate over 300 hyper-parameters sets using a Nvidia Titan-X (Pascal) for over 1,000 GPU hours. 

Taking into consideration of performance and efficiency trade-off, we select the DistilBERT model for our Transformer LM layer. DistilBERT is a compressed version of BERT Base through Knowledge Distillation. With utilization of only 50\% of the layers of BERT, DistilBERT performs 60\% faster while preserving 97\% of the capabilities in language understanding tasks. The optimal set of hyper-parameters for DistilBERT includes an AdamW~\cite{loshchilov2017adamw} optimizer, and initial learning rate of 5e-5, and a batch size of 32.

\textbf{Output Layer}. The output layer of \textit{DistilBERT} LM is a set of 768-d vectors led by the class header vector. Since we are conducting classification tasks, only the \textit{[CLS]} token vector is used as the aggregate sequence representation for classification with a linear classifier. The remainder of the output vectors are processed into \textit{Crisis2Vec} embeddings using Mean-Pooling operation.

\subsection{Crisis2Vec}

As discussed in Section~\ref{sec:CrisisBERT}, \textit{Crisis2Vec} embedding is a byproduct of \textit{CrisisBERT}, where the embeddings are constructed based on a pre-trained BERT model, and subsequently fine-tuned with three corpora of crisis-related tweets~\cite{olteanu2014crisislex, olteanu2015expect, zubiaga2016analysing} to be domain-specific for crisis-related tweet representation. 

\textit{Crisis2Vec} leverages the advantages of Transformers, including (1) leveraging a self-attention mechanism to incorporate sentence-level context bidirectionally, (2) leveraging both word-level and positional information to create contextual representation of words, and (3) taking advantage of the pre-trained models on large relevant corpora. 

To the best of our knowledge, we are the first who propose a document-level contextual embedding approach for crisis-related document representation. Upon convergence, we construct the fixed-length tweet vector using a MEAN-Pooling strategy~\cite{reimers2019sentence}, where we compute the mean of all output vectors, as illustrated in Algorithm ~\ref{alg:c2v}. 

\begin{algorithm}
\SetAlgoLined

\KwIn{$Tweet$ as a sequence of $Word$s}
\KwOut{$Tweet\_Embedding$}
$Initialize~~Token\_Vector$\;
\For{$word$ in $Tweet$} {
     $Word\_Token~\leftarrow~Embed(Word)$\;
     $Append~~Token\_Vector~\leftarrow~Word\_Token$\;
}
$Output\_Vectors~\leftarrow~CrisisBERT_{LM}(Token\_Vector)$\;
$Tweet\_Embedding~\leftarrow~Pool_{Mean}(Output\_Vectors)$\;
\KwRet{$Tweet\_Embedding$}\;

\caption{\textit{Crisis2Vec} with Mean Pooling}
\label{alg:c2v}
\end{algorithm}

\section{Crisis Classification}

In this work, we conduct two crisis classification tasks, namely \textit{Crisis Detection} and \textit{Crisis Recognition}. We formulate the \textit{Crisis Detection} task as a binary classification model that identifies if a tweet is relevant to a crisis-related event. The \textit{Crisis Recognition} task on the other hand extends the problem into multi-class classification, where the output is a probability vector that indicate the likelihood of a tweet indicating specific events. Both tasks are modelled as Sequence Classification problems that are formally defined below.

\subsection{Crisis Detection}

We define the Crisis Detection task $D= (S, \Phi)$, which is specified by $S=\{s_1, ..., s_n\}$ a finite sample space of tweets with size $n$. Each sample $s_i$ is a sequence of tokens at T time steps, i.e., $s_i=\{s_i^1, ..., s_i^T\}$. $\Phi$ denotes the set of labels that has the same sequence as the sample set, $\Phi=\{\phi_1, ..., \phi_n\}$ and $\phi_i\in\{0, 1\}$ where $\phi_i=1$ indicates that sample $s_i$ is relevant to crisis, and $\phi_i=0$ indicates otherwise. A deterministic classifier $C_D: S \rightarrow \phi$ specifies the mapping from sample tweets to their flags. 

Our objective is to train a crisis detector using the provided tweets and labels that minimizes the differences between predicted labels and true labels, i.e.,
\begin{equation}
  min.~\mathbb{J}_D(\Phi, C_D(S))
\end{equation}
where $\mathbb{J}_D$ denotes some cost function.

\subsection{Crisis Recognition}
Similarly, we define a Crisis Recognition task $R = (S, L)$, where sample space $S$ is identical to that in Crisis Detection. $L$ denotes a sequence of multi-class labels that have the same sequence as $S$, i.e., $L=\{l_1, ..., l_n\}$, and $l_i\in{\mathbb{R}^m}$ for $m$ number of classes. A deterministic classifier $C_R: S \rightarrow L$ specifies the mapping from the sample tweets to the crisis classes.

The objective of the crisis classification tasks is to train a sequence classifier using the provided tweets and labels that minimizes the differences between predicted labels and true labels, i.e., 
\begin{equation}
  min.~\mathbb{J}_R(L, C_R(S))
\end{equation}
where $\mathbb{J}_R$ denotes some cost function for classifier $C_R$.

\begin{table}[th] 
\resizebox{\textwidth}{!}{
\begin{tabular}{cccc|cccc}
\hline
Label          & Crisis event                      & \# Data Points           & Dataset    & Label          & Crisis event                      & \# Data Points           & Dataset  \\ \hline 
1              & 2012\_Sandy\_Hurricane            & 6318                     & C6            & 2              & 2013\_Alberta\_Floods             & 5189                     & C6            \\
3              & 2013\_Boston\_Bombings            & 6577                     & C6        & 4              & 2013\_Oklahoma\_Tornado           & 4827                     & C6            \\
5              & 2013\_Queensland\_Floods          & 6333                     & C6        & 6              & 2013\_West\_Texas\_Explosion      & 6157                     & C6            \\\hline
7              & sydneysiege                       & 837                      & C8        & 8              & charliehebdo                      & 903                      & C8            \\
9              & ferguson                          & 859                      & C8        & 10             & germanwings-crash                 & 248                      & C8            \\
11             & putinmissing                      & 59                       & C8        & 12             & ottawashooting                    & 639                      & C8            \\
13             & ebola-essien                      & 21                       & C8        & 14             & prince-toronto                    & 93                       & C8            \\ \hline
15             & 2012\_Colorado\_wildfires         & 953                      & C26       & 16             & 2012\_Costa\_Rica\_earthquake     & 909                      & C26           \\
17             & 2012\_Guatemala\_earthquake       & 940                      & C26       & 18             & 2012\_Italy\_earthquakes          & 940                      & C26           \\
19             & 2012\_Philipinnes\_floods         & 906                      & C26       & 20             & 2012\_Typhoon\_Pablo              & 907                      & C26           \\
21             & 2012\_Venezuela\_refinery         & 939                      & C26       & 22             & 2013\_Australia\_bushfire         & 949                      & C26           \\
23             & 2013\_Bohol\_earthquake           & 969                      & C26       & 24             & 2013\_Brazil\_nightclub\_fire     & 952                      & C26           \\
25             & 2013\_Colorado\_floods            & 925                      & C26       & 26             & 2013\_Glasgow\_helicopter\_crash  & 918                      & C26           \\
27             & 2013\_LA\_airport\_shootings      & 912                      & C26       & 28             & 2013\_Lac\_Megantic\_train\_crash & 966                      & C26           \\
29             & 2013\_Manila\_floods              & 921                      & C26       & 30             & 2013\_NY\_train\_crash            & 999                      & C26           \\
31             & 2013\_Russia\_meteor              & 1133                     & C26       & 32             & 2013\_Sardinia\_floods            & 926                      & C26           \\
33             & 2013\_Savar\_building\_collapse   & 911                      & C26       & 34             & 2013\_Singapore\_haze             & 933                      & C26           \\
35             & 2013\_Spain\_train\_crash         & 991                      & C26       & 36             & 2013\_Typhoon\_Yolanda            & 940                      & C26           \\ \hline
\end{tabular}}
\caption{Classes Description}
\label{tab:classes-desc}
\end{table}

\section{Experiments and Results}
In this section, we discuss the experiments performed and their results in order to propose a highly effective and efficient approach for text classification. 


\subsection{Datasets}

Three datasets of labelled crisis-related tweets~\cite{olteanu2014crisislex, olteanu2015expect, zubiaga2016analysing} are used to conduct crisis classification tasks and evaluate the proposed methods against benchmarks. In total, these datasets consist of close to 8 million tweets, where overall 91.6k are labelled. These data sets are in the form of: (1) 60k labelled tweets on 6 crises\cite{olteanu2014crisislex}, (2) 3.6k labelled tweets for 8 crises \cite{zubiaga2016analysing}, and (3) 27.9k labelled tweets for 26 crises \cite{olteanu2015expect}. Table~\ref{tab:classes-desc} describes more detail about each dataset and their respective classes.

For our experimental evaluation, the 91.6k labelled crisis-related tweets are organized into two datasets, annotated as C6 and C36. In particular, C6 consists of 60k tweets from 6 classes of crises, whereas C36 comprises all 91.6k tweets in 36 classes. Both datasets are split into training, validation and test sets that consist of 90\%, 5\% and 5\% of the original sets, respectively. Table \ref{tab:data-table} describes the statistics of the split sets. 

\begin{table}[h]
\centering
\begin{tabular}{cccccc}
\hline
Dataset                      &  \# Data Points &  \# Classes & Train & Test & Split \\ \hline
C6          & 60k                     & 6 + 1   & 54072 & 3004 & 3004          \\
C36         & 91.6k                   & 36 + 1   & 82506 & 4584 & 4584          \\ \hline
\end{tabular}
\caption{Dataset Description}
\label{tab:data-table}
\end{table}

\subsection{Proposed Models}

\textbf{CrisisBERT}. We evaluate the performance of \textit{CrisisBERT} against multiple benchmarks, which comprise recently proposed crisis classification models in the literature. These works include linear classifiers, such as Logistic Regression (LR), Support Vector Machine (SVM) and Naive Bayes~\cite{manna2019effectiveness}, and non-linear neural networks, such as Convolutional Neural Network (CNN)~\cite{kersten2019robust} and Long Short-Term Memory~\cite{kumar2020deep}. 

Furthermore, we investigate the robustness of \textit{CrisisBERT} for both detection and recognition tasks. This is achieved by extending the experiments from C6 to C36, which comprise 6 and 36 classes respectively, but with only 51.4\% additional data points. We evaluate the robustness of the proposed models against benchmarks by observing the compromise in robustness performance, while realizing the drastically improved classification performance.

As described in Section~\ref{sec:CrisisBERT}, we use the optimal set of hyper-parameters for \textit{CrisisBERT} in the experiments, which include the use of a BERT model with distillation (i.e. DistilBERT), an AdamW~\cite{loshchilov2017adamw} optimizer, an initial learning rate of 5e-5, a batch size of 32, and a word dropout rate of 0.25.

\textbf{Crisis2Vec}. To evaluate \textit{Crisis2Vec}, we choose the two classifiers with the aim to represent both traditional Machine Learning approaches and the NN approaches. The two selected models are: (1) a linear Logistic Regression model, denoted as $LR_{c2v}$, and (2) a non-linear LSTM model, denoted as $LSTM_{c2v}$. We evaluate the performance of \textit{Crisis2Vec} with the two models by replacing the original embedding to \textit{Crisis2Vec}, ceteris paribus.

\subsection{Metrics}
We use two common evaluation metrics, namely Accuracy and F1 score, which are functions of True-Positive (TP), False-Positive (FP), True-Negative (TN) and False-Negative (FN) predictions. Accuracy is calculated by:

\begin{equation}
Accuracy = \frac{TP+TN}{TP+TN+FP+FN}
\end{equation}

For a F1-score of multiple classes, we calculate the unweighted mean for each label, i.e., for $n$ classes of labels as:

\begin{equation}
F1 = \frac{2* Recall * Precision} {Recall + Precision}
\end{equation}
where 
\begin{equation}
 Precision = \frac{TP}{TP+FP}
\end{equation}
and
\begin{equation}
 Recall = \frac{TP}{TP+FN}
\end{equation}

\subsection{Benchmark Algorithms}

We select and implement several crisis classifiers proposed in recent works to serve as benchmarks for evaluating our proposed methods. Concretely, we compare \textit{CrisisBERT} with the following models:

\begin{itemize}
    \item $LR_{w2v}$: Logistic regression model with \textit{Word2Vec} embedding pre-trained on Google News Corpus~\cite{manna2019effectiveness}
    \item $SVM_{w2v}$: Support Vector Machine model with \textit{Word2Vec} embedding pre-trained on Google News Corpus ~\cite{manna2019effectiveness}
    \item $NB_{w2v}$: Naive Bayes model assuming Gaussian distribution for features with \textit{Word2Vec} embedding pre-trained on Google News Corpus~\cite{manna2019effectiveness}
    \item $CNN_{gv}$: Convolutional Neural Network model with 2 convolutional layers of 128 hidden units, kernel size of 3, pool size of 2, 250 filters, and \textit{GloVe} for word embedding~\cite{kersten2019robust}
    \item $LSTM_{w2v}$: Long Short-Term Memory model with 2 layers of 30 hidden states and a \textit{Word2Vec}-based Crisis Embedding~\cite{kumar2020deep}
\end{itemize}

\begin{table*}[h]
\centering
\resizebox{.85\textwidth}{!}{%
\begin{tabular}{lcccccccc}
\hline
\multirow{2}{*}{Models} & \multicolumn{2}{c}{Detection-C6} & \multicolumn{2}{c}{Detection-C36} & \multicolumn{2}{c}{Recognition-C6} & \multicolumn{2}{c}{Recognition-C36} \\ \cline{2-9} 
 & F1  & Accuracy  & F1  & Accuracy  & F1  & Accuracy  & F1  & Accuracy  \\ \hline
\multicolumn{9}{l}{\textit{\textbf{Proposed}}} \\
$CrisisBERT$ & \textbf{95.5} & \textbf{95.6} & \textbf{94.2} & \textbf{94.7} & \textbf{98.7} & \textbf{98.6} & \textbf{97.1} & \textbf{97.9}  \\ 
$LSTM_{c2v}$ & 95.1 & 95.1 & 92.8 & 93.5 & 97.5 & 97.5 & 88.0 & 95.6  \\ 
$LR_{c2v}$ & 93.2 & 93.2 & 89.0  & 90.1  & 93.6 & 93.7 & 85.1 & 90.9 \\ \hline
\multicolumn{9}{l}{\textit{\textbf{Benchmark}}} \\
$LR_{w2v}$~\cite{manna2019effectiveness} & 91.5 & 91.6 & 85.3 & 86.6 & 87.3 & 88.5 & 72.1 & 82.3 \\
$SVM_{w2v}$~\cite{manna2019effectiveness} & 91.3 & 91.4 & 85.0 & 86.3 & 86.6 & 87.9 & 71.6 & 81.9  \\
$NB_{w2v}$~\cite{manna2019effectiveness} & 86.8 & 86.8 & 82.4 & 83.4 & 78.8 & 80.5 & 47.6 & 63.5  \\
$CNN_{gv}$~\cite{kersten2019robust} & 91.2 & 91.3 & 91.1 & 91.2 & 90.5 & 90.4 & 23.3 & 64.4 \\
$LSTM_{w2v}$~\cite{kumar2020deep} & 91.7 & 91.7 &  88.0  & 89.3   & 87.3 &  87.3  & 58.9 & 72.3 \\
\hline
\end{tabular}}
\caption{Experimental results of Crisis Classification tasks on C6 and C36 datasets for proposed models and benchmarks, where best performers are emphasized. Results show that \textit{CrisisBERT} records highest performance across all tasks. }
\label{tab:exp-results}
\end{table*}

\subsection{Results}

Overall, the experimental results show that both proposed models achieve significant improvement on performance and robustness over benchmarks across all tasks. The experimental results for \textit{CrisisBERT} and \textit{Crisis2Vec} are tabulated in Table~\ref{tab:exp-results}. 

\textbf{Crisis Detection}. For Crisis Detection tasks on C6 dataset, \textit{CrisisBERT} achieves a 95.5\% F1-score and 95.6\% Accuracy, which exceeds previous best model results, namely \textit{LSTM} with pre-trained \textit{Word2Vec} embeddings, by 3.8\% and 3.8\% respectively. In terms of embedding, \textit{LSTM} with \textit{Crisis2Vec} records 95.1\% for both F1-score and Accuracy, which shows 3.4\% improvement over \textit{LSTM} with \textit{Word2Vec}. Similarly, LR with \textit{Crisis2Vec} records 93.2\% for both F1-score and Accuracy, which shows 1.6\% improvement over \textit{LR} with \textit{Word2Vec}.

For Crisis Detection tasks on C36 dataset, \textit{CrisisBERT} achieves 94.2\% F1-score and 94.7\% Accuracy, which exceeds previous best model results, namely \textit{CNN} with pre-trained \textit{GloVe} embedding, by 2.7\% and 2.1\% respectively. In terms of embeddings, \textit{LSTM} with \textit{Crisis2Vec} records 92.8\% F1-score and 93.5\% Accuracy, which exceed \textit{LSTM} with \textit{Word2Vec} by 4.8\% and 4.2\%. Similarly, \textit{LR} with \textit{Crisis2Vec} records 89.0\% F1-score and 90.1\% Accuracy, which exceed \textit{LR} with \textit{Word2Vec} by 3.7\% and 3.5\% respectively. 

\textbf{Crisis Recognition}. For Crisis Recognition tasks on C6 dataset, \textit{CrisisBERT} achieves 98.7\% F1-score and 98.6\% Accuracy, which exceeds previous best model results, namely \textit{CNN} with pre-trained \textit{GloVe} embeddings, by 8.2\% for both F1-score and Accuracy. In terms of embedding, \textit{LSTM} with \textit{Crisis2Vec} records 97.5\% for both F1-score and Accuracy, which shows a significant improvement of 10.2\% over \textit{LSTM} with \textit{Word2Vec}. Similarly, LR with \textit{Crisis2Vec} records 93.6\% F1-score and 93.7\% Accuracy, which shows 6.3\% and 5.2\% improvement respectively over \textit{LR} with \textit{Word2Vec}.

For Crisis Detection tasks on C36 dataset, \textit{CrisisBERT} achieves 97.1\% F1-score and 97.9\% Accuracy, which significantly exceeds previous best model results, namely \textit{LR} with pre-trained \textit{Word2Vec} embedding, by 25.0\% and 15.6\% respectively. In terms of embeddings, \textit{LSTM} with \textit{Crisis2Vec} records 88.0\% F1-score and 95.6\% Accuracy, which significantly exceed \textit{LSTM} with \textit{Word2Vec} by 29.1\% and 23.3\%. Similarly, \textit{LR} with \textit{Crisis2Vec} records 85.1\% F1-score and 90.9\% Accuracy, which significantly exceed \textit{LR} with \textit{Word2Vec} by 13.0\% and 8.6\% respectively.

\textbf{Robustness}. Comparing Crisis Detection task between C6 and C36, \textit{CrisisBERT} shows 1.7\% and 1.3\% decline for F1-score and Accuracy, which is much better than most benchmarks, i.e., from 1.7\% to 6.3\%, except \textit{CNN}. However, when we compare the more challenging Crisis Recognition tasks between C6 and C36, the performance of \textit{CrisisBERT} compromises marginally, i.e., 1.6\% for F1-score and 0.7\% for Accuracy. On the contrary, all benchmark models record significant decline, i.e. from 6.0\% to 67.2\%. 

\textbf{Discussion}. Based on experimental results discussed above, we observe that: (1) \textit{CrisisBERT}'s performance exceeds state-of-the-art performance for both detection and recognition tasks, with up to 8.2\% and 25.0\% respectively, (2) \textit{CrisisBERT} demonstrates higher robustness with marginal decline for performance (i.e. less than 1.7\% in F1-score and Accuracy), and (3) \textit{Crisis2Vec} shows superior performance as compared to conventional \textit{Word2Vec} embeddings, for both \textit{LR} and \textit{LSTM} models across all experiments.

\section{Related Work}

\subsection{Crisis Classification}
Event detection from tweets has received significant attention in research in order to analyze crisis-related messages for better disaster management and increasing situational awareness~\cite{imran2015processing, sakaki2010earthquake,said2019natural,snyder2019situational}. Parilla-Ferrer \textit{et al.} proposed automatic binary classification of informativeness using Naive Bayes and Support Vector Machine (SVM)~\cite{parilla2014automatic}. Stowe \textit{et al.} presented an annotation scheme for tweets to classify relevancy and six ~\cite{stowe2016identifying}. Furthermore, use of pre-trained word2vec reportedly improved SVM for Crisis classification~\cite{mikolov2013distributed}. Lim \textit{et al.} proposed CLUSTOP algorithm utilizing the community detection approach for automatic topic modelling~\cite{lim2017clustop}. Pedrood \textit{et al.} proposed to transfer-learn classification of one event to the other using a sparse coding model~\cite{pedrood2018mining}, though the scope was only limited to only two events, i.e. Hurricane Sandy (2012) and Supertyphoon Yolanda (2013).

A substantial number of works focusses  on usign Neural Networks (NN) with word embeddings for crisis-related data classification. Manna \textit{et al.} \cite{manna2019effectiveness} compared NN models with conventional ML classifiers~\cite{manna2019effectiveness}. ALRashdi and O’Keefe investigated and showed good performance for two deep learning architectures, namely Bidirectional Long Short-Term Memory (BiLSTM) and Convolutional Neural Networks (CNN) with domain-specific GloVe embeddings~\cite{alrashdi2019deep}. However, the study had yet to validate the relevance of model on a different crisis type. Nguyen \textit{et al.} applied CNN to classify information types using Google News and domain-specific embeddings\cite{nguyen2017robust}. Kersten \textit{et al.} \cite{kersten2019robust} implemented a parallel CNN to detect two disasters, namely hurricanes and floods, which reported a F1-score of 0.83. The CNN architecture was proposed earlier by Kim \textit{et al.} ~\cite{kim2014convolutional}. 

\subsection{Crisis Embedding}

Word-level Embeddings such as \textit{Word2Vec}~\cite{bengio2003neural} and GloVe~\cite{pennington2014glove} are commonly used to form the basis of Crisis Embedding~\cite{nguyen2017robust, nguyen2016applications} in various crisis classification works to improve model performance. For context, \textit{Word2Vec} uses a Neural Network Language Model (NNLM) that is able to represent latent information on the word level. \textit{GloVe} achieved better results with a simpler approach, constructing global vectors to represent contextual knowledge of the vocabulary. 

More recently, a series of high quality embedding models, such as FastText~\cite{bojanowski2017fasttext} and Flair~\cite{akbik2018contextual}, are proposed and reported to have improved state of the art for multiple NLP tasks. Both word-level contextualization and character-level features are commonly used for these works. Pre-trained models on large corpora of news and tweets collections are also made publicly available to assist in downstream tasks. Furthermore, Transformer-based models are proposed to conduct sentence-level embedding tasks~\cite{reimers2019sentence}. 

\section{Conclusion}

Social media such as Twitter has become a hub of crowd generated information for early crisis detection and recognition tasks. In this work, we present a transformer-based crisis classification model \textit{CrisisBERT}, and a contextual crisis-related tweet embedding model \textit{Crisis2Vec}. We examine the performance and robustness of the proposed models by conducting experiments with three datasets and two crisis classification tasks. Experimental results show that \textit{CrisisBERT} improves state of the art for both detection and recognition class, and further demonstrates robustness by extending from 6 classes to 36 classes, with only 51.4\% additioanl data points. Finally, our experiments with two classification models show that \textit{Crisis2Vec} enhances classification performance as compared to \textit{Word2Vec} embeddings, which is commonly used in prior works.

\section{Acknowledgement}

This research is funded in part by the Singapore University of Technology and Design under grant SRG-ISTD-2018-140.

\bibliographystyle{unsrt}  
\bibliography{crisisBERT}  

\end{document}